\documentclass[sigconf,authorversion]{acmart}
\AtBeginDocument{%
  \providecommand\BibTeX{{%
    \normalfont B\kern-0.5em{\scshape i\kern-0.25em b}\kern-0.8em\TeX}}}

\usepackage{natbib}
\usepackage{enumitem}
\usepackage{wrapfig}
\usepackage{color,soul}

\begin{document}

\title{A Computational Inflection for Scientific Discovery}


\author{Tom Hope}
\email{tomh@allenai.org}
\affiliation{
  \institution{The Allen Institute for AI}
  \institution{The Hebrew University of Jerusalem}
  \country{}
}

\author{Doug Downey}
\email{dougd@allenai.org}
\affiliation{
  \institution{The Allen Institute for AI}
  \institution{Northwestern University}
  \country{}
}

\author{Oren Etzioni}
\email{oren@allenai.org}
\affiliation{
  \institution{The Allen Institute for AI}
  \country{}
}

\author{Daniel S. Weld}
\email{danw@allenai.org}
\affiliation{
  \institution{The Allen Institute for AI}
  \institution{The University of Washington}
  \country{}
}

\author{Eric Horvitz}
\email{horvitz@microsoft.com}
\affiliation{
   \institution{Office of the Chief Scientific Officer}
   \institution{Microsoft}
  \country{}
}






\renewcommand\footnotetextcopyrightpermission[1]{}
\pagestyle{plain}
\maketitle

\section{Abstract}
We stand at the foot of a significant inflection in the trajectory of scientific discovery.
As society continues on its fast-paced digital transformation, so does humankind's collective scientific knowledge and discourse. We now read and write papers in digitized form, and a great deal of the formal and informal processes of science are captured digitally---including papers, preprints and books, 
code and datasets, conference presentations, and interactions in social networks and collaboration and communication platforms. The transition has led to the creation and growth of a tremendous amount of information---much of which is available for public access---opening exciting opportunities for computational models and systems that analyze and harness it. In parallel, exponential growth in data processing power has fueled remarkable advances in artificial intelligence, including large neural language models capable of learning powerful representations from unstructured text. Dramatic changes in scientific communication---such as the advent of the first scientific journal in the 17$^{\text{th}}$ century---have historically catalyzed revolutions in scientific thought. The confluence of societal and computational trends suggests that computer science is poised to ignite a revolution in the scientific process itself.

\section{Introduction}

At the heart of the scientific process, a basic behavior has remained unchanged for hundreds of years: We build on \emph{existing} ideas for forming \emph{new} ideas. When faced with a new question or problem, we leverage knowledge from accumulated learnings and from external sources and perform synthesis and reasoning to generate insights, answers, and directions.
 But the last few decades have brought changes.
 The explosion of digital information and steep acceleration in the production of scientific data, results and publications---with over one million papers added every year to the PubMed biomedical index alone---stand in stark contrast to the constancy of human cognitive capacity. While scientific knowledge, discourse, and the larger scientific ecosystem are expanding with rapidity, our human minds have remained static, with severe limitations on the capacity for finding, assimilating and manipulating information. Herbert Simon's reflection that ``…a wealth of information creates a poverty of attention,'' aptly describes the limited attention of researchers in the modern scientific ecosystem. 
 Even within narrow areas of interest, there is a vast space of potential directions to explore---while the keyhole of cognition admits only a tiny fraction of the broad landscape of information and deliberates over small slices of possibility. The way we search through and reflect about information across the vast space---the areas we select to explore, and how we explore them---is hindered by cognitive biases \cite{nuzzo2015scientists} and lacks principled and scalable tools for guiding our attention \cite{AttentionBiases}. ``Unknowns'' are not just holes in science, but important gaps in personal knowledge about the broader knowns across the sciences. 

We thus face an imbalance between the treasure trove of scholarly information and our limited ability to reach into it. Despite technological advances, we require new paradigms and capabilities to address this widening gap. We see promise in developing new foundational capabilities that address the cognitive bottleneck, aimed at extending human performance on core tasks of research---e.g., keeping abreast with developments, forming and prioritizing ideas, conducting experiments, reading and understanding papers (see Table \ref{tab:tasks}). We focus on a research agenda we call \emph{task-guided scientific knowledge retrieval}, in which systems counter humans' bounded capacity by ingesting corpora of scientific knowledge and retrieving inspirations, explanations, solutions and evidence synthesized to directly serve task-specific utility. We present key concepts of task-guided scientific knowledge retrieval, including work on prototypes that highlight the promise of the direction and bring into focus concrete steps forward for novel representations, tools, and services. In Section \ref{sec:taskguided} we review systems 
that help researchers discover novel perspectives and inspirations \cite{portenoy2021bridger,hope2020scisight, hope2021scaling, hope2017accelerating}, help guide the attention of researchers toward opportunity areas rife with uncertainties and unknowns \cite{lahav2021search,AttentionBiases}, and models that leverage retrieval and synthesis of scientific knowledge as part of machine learning and prediction \cite{naik2021literature, gil2022will}. We conclude in Section \ref{sec:challenges} with a discussion of opportunities ahead with computational approaches that have the potential to revolutionize science. 

To set the stage, in the following section we begin by discussing some fundamental concepts and background for our research agenda.

\begin{table*}
{
\begin{tabular}{p{0.27\linewidth}|p{0.69\linewidth}}
    \hline
    \textbf{Task/Activity} & \textbf{Description}
    \\
    \hline
    \parbox[t]{\linewidth}{\emph{Attention to areas of interest}}
    &\parbox[t]{\linewidth}{A background process of keeping track of latest developments in relevant scientific communities. Involves applying selective attention, perceiving relevance and utility.}
    \\
    \hline    
    \parbox[t]{\linewidth}{\emph{Problem identification \&  prioritization}}
    &\parbox[t]{\linewidth}{Identifying new research questions and deciding on which ones to work. Involves factors such as subjective preferences and assessment of feasibility.}
    \\
    \hline
    \parbox[t]{\linewidth}{\emph{Forming directions}}
    &\parbox[t]{\linewidth}{Given a problem/question, forming ideas to address it. Involves cognitive processes such as constructing mental models of a problem, problem reformulation, abstraction and decomposition, adaptation of relevant knowledge to new scenarios, and assessing likelihood of success.}
        \\
        \hline
    \parbox[t]{\linewidth}{\emph{Literature search \&  review}}
    &\parbox[t]{\linewidth}{Accessing and ingesting knowledge in the literature. Involves many processes such as query formulation, skimming and assessing relevance, positioning ideas with relations and contrasts to existing work, and reading and summarization strategies.}
        \\
        \hline
    \parbox[t]{\linewidth}{\emph{Learning, understanding, sense-making}}
    &\parbox[t]{\linewidth}{The cognitive processes and activities involved in assimilating new knowledge and concepts, and making sense of complex scientific information spaces.}
    \\
        \hline
    \parbox[t]{\linewidth}{\emph{Experimentation, analysis, action}}
    &\parbox[t]{\linewidth}{A broad category referring to the many processes and activities involved in formulating and conducting experiments (e.g., planning data collection and measurements), performing analyses (e.g., understanding a set of data points, modeling and extrapolation, prediction, evaluation), and producing artifacts, techniques, theories, decisions, policies, actions.}
    \\
    \hline
    \parbox[t]{\linewidth}{\emph{Research communication}}
    &\parbox[t]{\linewidth}{Writing research documents (papers, proposals, analyses), communicating with peers (feedback and review, collaboration, presentation).}
   \\
    \hline
\end{tabular}
}
\
\caption{Research may be decomposed into salient tasks that are prime targets for computational augmentation (§~\protect\ref{sec:taskguided}).}
\label{tab:tasks}
\end{table*}

\section{Human-Centric Perspective}
\label{sec:scienv}

Extraordinary developments at the convergence of AI and scientific discovery have emerged in specific areas, including new kinds of analytical tools, with the prominent example of AlphaFold, which harnesses deep neural models to dramatically improve the prediction of protein structure from amino acid sequence information \cite{jumper2021highly}. Large language models (LLMs) have very recently made stellar progress in the ability to reason about complex tasks, including in the medical domain \cite{nori2023capabilities}. The most advanced LLM at present---emerging before the ink has dried on this paper---is GPT-4, which has exhibited jaw-dropping skill at handling clinical questions, mathematical problems and computer coding tasks \cite{bubeck2023sparks}.

\begin{figure}
    \centering
    \includegraphics[width=2.5in]{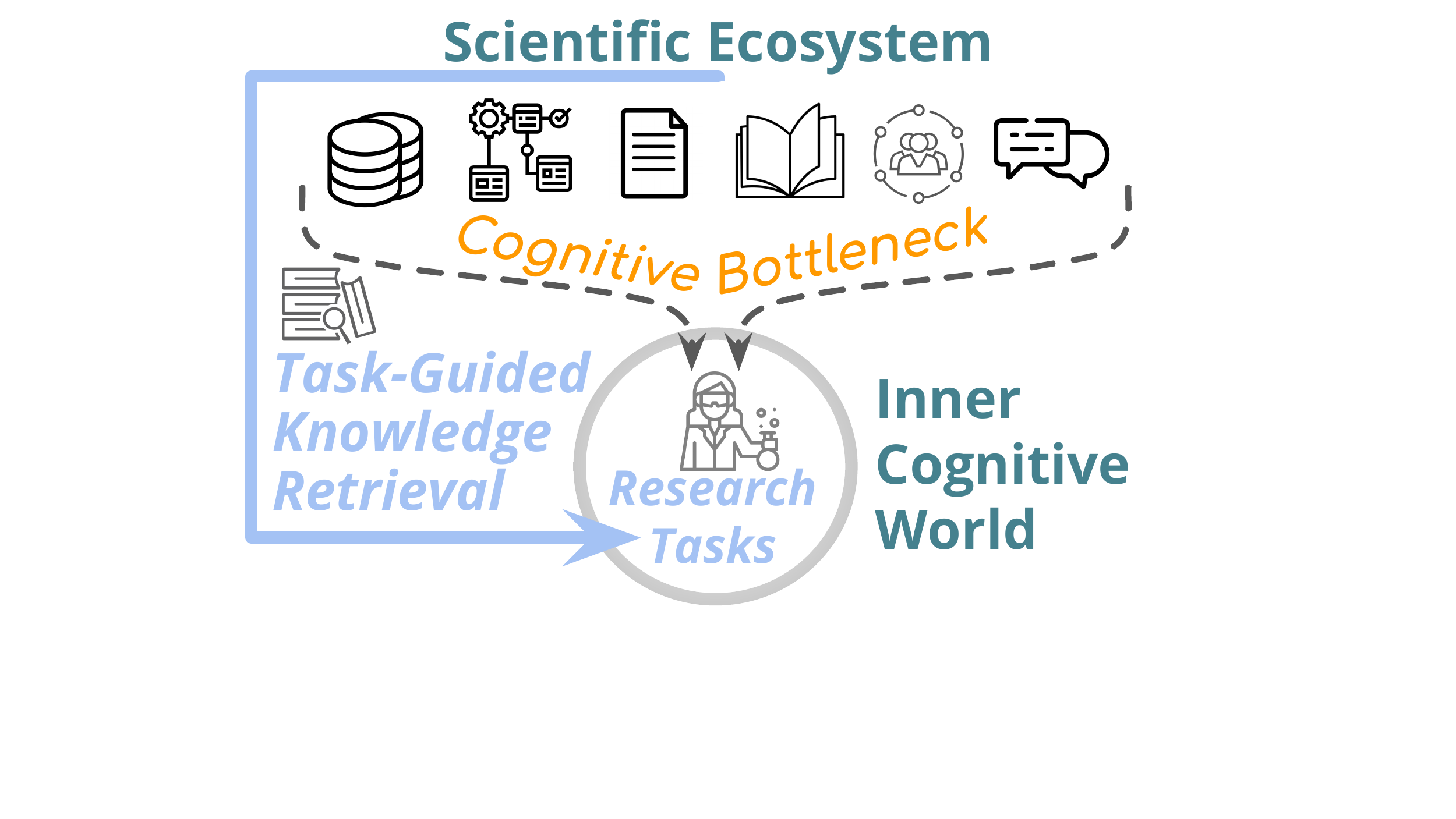}
    \vspace*{-.2in}
    \caption{
    Information flows from the outer world into the inner cognitive world of researchers, constrained by cognitive capacity and biases. 
    We see opportunities to support researchers by retrieving knowledge that helps with tasks across multiple phases of the scientific process (Table \ref{tab:tasks}).}
    \label{fig:innerouter}
\end{figure}

{We view these developments as tremendous research opportunities for building computational approaches that accelerate scientific discovery}. We take a \emph{human-centered}, cognitive perspective: augmenting researchers by taking into account the diversity of tasks, contexts, and cognitive processes involved in consuming and producing scientific knowledge. Collectively, we refer to these as the \emph{inner cognitive world} of a researcher\footnote{We use the term \emph{researcher} to include also practitioners in science-driven areas, such as medical doctors and technological engineers, who require deep scientific knowledge.} (see Figure \ref{fig:innerouter}). The researcher interacts with the scientific ecosystem---literature, resources, discussions---in order to inform decisions and actions. Researchers have different uses for scholarly information, depending on the task at hand and the stage of exploration (see Table \ref{tab:tasks} and discussion in Section \ref{sec:taskguided}). 
We pursue a research agenda around assisting researchers in their tasks, guided by two main desiderata:

\begin{enumerate}[wide]
    \item Systems for augmenting human capabilities in the sciences need to enhance the \emph{effective flow of knowledge from the outer world of scientific information and discourse to the researcher's inner cognitive world}---countering humans' bounded capacity by retrieving and synthesizing information targeted to enhance performance on tasks. Achieving this goal requires methods that build and leverage rich representations of scientific content and that can align computational representations with human representations, in the context of specific tasks and backgrounds of researchers.
    \item Research on such systems should be rooted in \emph{conceptual models of the inner cognitive world of a researcher}. Shining a spotlight on this inner world brings numerous factors and questions to the fore. How do researchers form ideas? How do they decide which problems to look into? How do they find and assimilate new information in the process of making decisions? What cognitive representations and bottlenecks are involved? What computing services would best augment these processes?

\end{enumerate}

\textbf{Background and Related Themes.} We leverage research in natural language processing, information retrieval, data mining and human-computer interaction and draw concepts from multiple disciplines. For example, efforts in \emph{metascience} focus on sociological factors that influence the evolution of science \cite{kuhn1970structure}, e.g., analyses of {information silos} that impede mutual understanding and interaction \cite{vilhena2014finding}, analyses of macro-scale ramifications of the rapid growth in scholarly publications \cite{chu2021slowed}, and of current metrics for measuring impact \cite{garcia2021critical} --- work enabled by digitization of scholarly corpora (see Section \ref{subsec:outer}). Metascience research makes important observations about human biases (desideratum 2) but generally does not engage in building computational interventions to augment researchers (desideratum 1). Conversely, work in \emph{literature-based discovery} \cite{Swanson1986-SWAFOR} mines information from literature to generate new predictions (e.g., functions of materials or drug targets) but is typically done in isolation from cognitive considerations; however, these techniques have great promise in being used as part of human-augmentation systems (see Sections \ref{sec:taskguided}-\ref{sec:challenges}). Other work uses machines to automate aspects of science. Pioneering work from Herbert Simon and Pat Langley automated discovery of empirical laws from data, with models inspired by cognitive mechanisms of discovery (see Section \ref{subsec:inner}). More recent work has focused on developing \emph{robot scientists} \cite{king2004functional,pyzer2022accelerating} that run certain experiments in biology or chemistry---not only formulating hypotheses but ``closing the loop'' by automated tests in a physical laboratory---where robots may use narrow curated background knowledge (e.g., of a specific gene regulatory network) and machine learning to guide new experiments. Related work explores automating scientific data analysis \cite{gil2022will}, which we discuss in Section \ref{sec:taskguided} as a case of retrieval from scientific repositories to augment aspects of \emph{experimentation and analysis} (see Table \ref{tab:tasks}).

We now turn to a discussion of central concepts: the ecosystem of science, and the cognitive world. This presentation lays the foundations for our exposition of task-guided retrieval in Section \ref{sec:taskguided} and the research opportunities in Section \ref{sec:challenges}. 

\subsection{Outer World: Scientific Ecosystem} 
\label{subsec:outer}
We collectively name the scientific ecosystem and the digital representations of scientific knowledge as the \emph{outer world} (see Figure \ref{fig:innerouter}). 
The outer world is comprised of {scientific communities}, a complex and shifting web of peers, concepts, methodologies, problems and directions revolving around shared interests, understandings and paradigms. This ecosystem generates digital information---digital ``traces'' of scientific thought and behavior---lying at the center of our attention as computer scientists interested in boosting human capacity to ``reach into'' the  pool of scientific knowledge. This knowledge includes \emph{scholarly publications} that appear in journals, conference proceedings, and online preprint repositories. Online publications are a main case of \emph{digital research artifacts}; other examples of products of research include software, datasets, knowledge bases. Research artifacts are also associated typically with \emph{signals of quality and interest}, such as citations to a specific paper or downloads of a dataset. The specific context of why a paper or resource was cited or used is often reflected in natural language descriptions. Different types of signals include \emph{peer review} prior to publication (mostly not shared publicly), and \emph{social media discussions} such as on Twitter, which has become a major virtual platform for academic dissemination and conversation. Along with the trend in society, \emph{private communication channels} among researchers are also digital---emails, online calls and messages. Similarly, \emph{note taking and writing}---important activities across  the scientific workflow---are done in digital form. This information is siloed in different platforms under privacy restrictions, yet represents a treasure trove for tools for the augmentation of scientific reasoning and exploration.

\subsection{Inner World: Human Cognition in Science}
\label{subsec:inner}

The way researchers decide to interact with information in the outer world and the way they process and use this information is governed by a complex array of cognitive processes, personal knowledge and preferences, biases and limitations, which are only partially understood. We collectively name these the \emph{inner world}, and briefly discuss several salient aspects.

Early work in AI by Herbert Simon and Alan Newell and later efforts by Pat Langley and Paul Thagard focused on cognitive and computational aspects of problem solving, creativity, decision making and scientific reasoning and discovery, seeking algorithmic representations to help understand and mimic human intelligence \cite{langley1987scientific,thagard2012cognitive}. Cognitive mechanisms that play important roles in scientific discovery include inductive and abductive reasoning, mental modeling of problems and situations, abstraction, decomposition, reformulation, analogical transfer and recombination; for example, in analogical transfer, given a situation or problem being considered in our working memory, we retrieve from our long-term memory prior analogous problems or situations. 

This cognitive machinery powers humans' ingenuity. However, the human mind also has severe limitations---\emph{bounded rationality} in the words of Simon---that impede these powerful mechanisms. Our limitations and capabilities have been studied for over a hundred years with cognitive psychology. Our limitations manifest in bounded cognitive capacity and knowledge, and biases that govern our behaviors and preferences. These limitations are all tightly interrelated. The ability to generate ideas, for instance, directly relies on prior knowledge; but, when a large volume of information from the outer world of science is met by insufficient cognitive capacity for processing and assimilating it, the result is information overload---a ubiquitous hindrance for researchers \cite{portenoy2021bridger}. Information overload in science strains the attentional resources of researchers, and forces researchers to allocate attention to increasingly narrow areas. This effect, in turn, amplifies a host of biases which researchers, just like all humans, suffer from \cite{nuzzo2015scientists, AttentionBiases}.  For example, scientists can be limited by confirmation bias, aversion to information from novel domains, homophily, and fixation on specific directions and perspectives without consideration of alternative views \cite{nuzzo2015scientists,hope2021scaling}. More broadly, selection of directions and areas to work on is  a case of decision-making, and as such {personal preference} and {subjective utility} play fundamental roles. Our research decisions rely on subjective assessment of feasibility, long-term or short-term goals and interests, and even psychological factors (e.g., tendencies for risk aversion). These factors are of course also impacted by biases \cite{nuzzo2015scientists}.

Clearly, the inner world of researchers is dauntingly complex. However, in the next section, we present encouraging results of applying computational methods to augment cognition in the sciences, helping to mitigate biases and limitations and enabling researchers to make better use of their powerful creative mechanisms.

\section{Task-Guided Retrieval}

\label{sec:taskguided}

How might we widen and deepen the connection between the outer world of science and researchers' limited cognitive worlds? We see a key bridge and research opportunity with developing tools for scientific \emph{task-guided knowledge retrieval}. In this section, we discuss our vision and present initial work toward achieving it.

Drawing from discussions in literature on the process of scientific discovery, we enumerate in Table \ref{tab:tasks} salient scientific tasks and activities, such as \emph{problem identification, forming directions, learning, literature search and review, experimentation}. These tasks could benefit from augmentation of human capabilities but remain underexplored in computer science. Existing computational technologies for assisting humans in discovering scientific knowledge are underinvested in important aspects of the intricate cognitive processes and goal-oriented contexts involved in scholarly endeavors.

The dominant approach to information retrieval research and systems can be summarized as ``relevance first''---focusing on results that answer user queries as accurately as possible. Academic search engines assume users know what queries to explore and how to formulate them. For pinpointed literature search in familiar areas, this assumption \emph{may} often suffice; but a broad array of other scholarly tasks, such as ideation or learning about a new topic, are very much underserved \cite{hope2020scisight,hope2021mechanisms, lahav2021search, hope2021scaling, portenoy2021bridger}.  At the same time, many voices in the information retrieval community have discussed a different, broader view of \emph{utility-driven} search situated in a wider context of information seeking by users with specific intents and tasks \cite{shah2022situating}. Here, we adapt ideas and principles from this general paradigm. 

We envision methods for task-guided scientific knowledge retrieval: systems that retrieve and synthesize outer knowledge in a manner that directly serves a task-guided utility of a researcher, while taking into consideration the researcher's goals, state of inner knowledge, and preferences.

Consider the tasks in Table \ref{tab:tasks}. For researchers engaged in \emph{experimentation or analysis}, we envision systems that help users identify experiments and analyses in the literature to guide design choices and decisions. For researchers in early stages of \emph{selecting problems} to work on, we picture systems that support this decision with information from literature and online discussions, synthesized to obtain estimated impact and feasibility. As part of \emph{forming directions} to address a problem, systems will help users find inspirations for solutions. Researchers who are \emph{learning} about a new topic will be provided with retrieved texts and discussions that explain the topic in a manner personally tailored to personal knowledge. Importantly, task-guided knowledge retrieval follows the two desiderata introduced in Section \ref{sec:scienv}; namely, systems should enable users to find knowledge that directly assists them in core research tasks by augmenting their cognitive capacity and mitigating their biases, and computational representations and services should align with salient cognitive aspects of the inner world of researchers.

\subsection{Prototypes of Task-Guided Retrieval}

We present work on initial steps and prototypes, including representative work that we have done and the work of others, framed in alignment with task-guided knowledge retrieval and tasks enumerated in Table \ref{tab:tasks}. The main aim of this brief review is to stimulate discussion in the computer science community on tools for extending human capabilities in the sciences. Existing methods are far from able to realize our vision. For example, we see major challenges in representation and inferences about the inner world of knowledge and preferences, and aligning these with representations and inferences drawn from the outer world knowledge. Today's prototypes are limited examples of our vision, using very rough proxies of inner knowledge and interest based on papers and documents written or read by the user, or in some cases only a set of keywords. We discuss these research challenges and others in Section \ref{sec:challenges}.

\textbf{Forming Directions.} We have developed methods for helping researchers generate new directions. A fundamental pattern in the cognitive process of creativity involves detecting \emph{abstract connections} across ideas and transferring ideas from one problem to another \cite{thagard2012cognitive}. Grounded in this cognitive understanding, we have pursued several approaches for stimulating creativity powered by retrieving outer knowledge. We developed and studied a system named Bridger that connects researchers to peers who inspire novel directions for research \cite{portenoy2021bridger}. Bridger identifies matches among authors based on \emph{commonalities and contrasts}, identifying peers who are both relevant and novel---working on similar problems but using very different methods, potentially inspiring new solutions. By doing so, Bridger helps mitigate the cognitive bias of \emph{fixation} \cite{hope2021scaling}. 

In this setting, {inner knowledge} is represented as mentions of problems and methods extracted automatically from a researcher's papers and weighted by term frequency. The {outer knowledge} being retrieved takes the form of other authors in computer science, following the same representation.  For each retrieved author, the system displays  salient problems, methods and papers, ranked by measures of relevance to the user. In studies with CS researchers, we found that Bridger dramatically boosted creative search and inspiration over state-of-art neural models employed by the Semantic Scholar search engine, surfacing useful connections across diverse areas; for example, one researcher drew novel connections between
the mathematical area of graph theory and their own area of human-centered AI, by exploring a recommended author who applies graph theory to decision making. The studies also surfaced important challenges, discussed in Section \ref{sec:challenges}.
\begin{figure}[h]
    \centering
    \includegraphics[width=0.95\columnwidth]{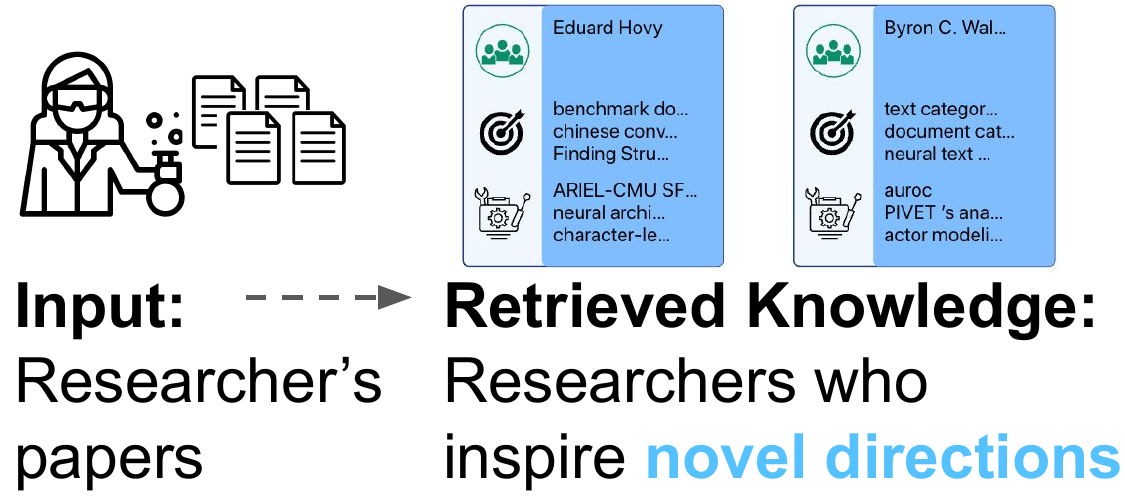}
    \caption{Matching researchers to authors with whom they are unfamiliar, to help in generating directions. Author cards show key problems and methods extracted from their papers.}
    \label{fig:bridger}
\end{figure}

We have also explored retrieving {outer knowledge} to enhance the human ability to find opportunities for analogical transfer \cite{hope2017accelerating,chan2018solvent}. Extensive work in cognitive studies has highlighted the human knack for ``analogical retrieval'' as a central function in creativity---bringing together structurally related ideas and adapting them to a task at hand \cite{thagard2012cognitive}. We developed a search method that enables researchers to search through a database of technological inventions and find mechanisms that can be transferred from distant domains to solve a given problem. Given as input from the user a textual description of an invention, we retrieve ideas (inventions, papers) that have partial structural similarity to the input (e.g., inventions with similar mechanisms), to facilitate discovery of analogical transfer opportunities. We found that the method could significantly boost measures of human creativity in ideation experiments, in which users were asked to formulate new ideas after viewing inspirations retrieved with our approach versus baseline information retrieval methods. For example, a biomechanical engineering lab working on polymer stretching/folding for creating novel structures found useful inspiration in a civil engineering paper on web crippling in steel beams---abstractly related to stretching and folding.

\begin{wrapfigure}{l}{0.45\columnwidth}
\includegraphics[width=0.45\columnwidth]{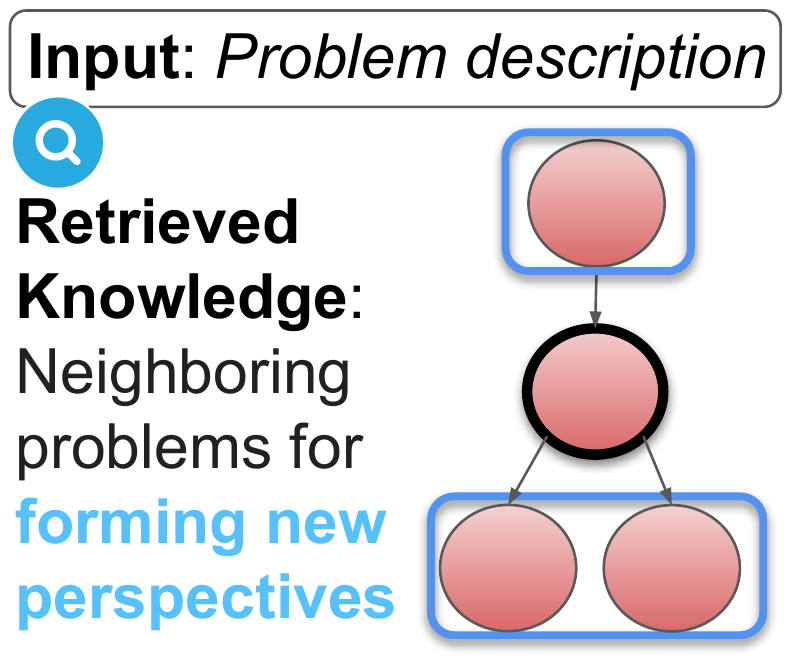}
    \caption{
    Using an extracted hierarchy of problems to retrieve new perspectives on a focal problem of interest.}
    \label{fig:problems}
\end{wrapfigure}

Innovation may also involve traversing multiple levels of abstraction around a problem to ``break out'' of fixation on the details of a specific problem by exploring novel perspectives. Given as input a problem description written by the user (as a proxy summary of the user's inner world of knowledge and purpose), we have pursued mechanisms that can retrieve diverse problem perspectives that are related to the focal problem, with the goal of inspiring new ideas for problem abstraction and reformulation \cite{hope2021scaling} (see Figure \ref{fig:problems}). Using NLP models to extract mentions of problems, we mine a corpus of technological invention texts to discover problems that often appear together, and use this information to form a hierarchical problem graph that supports automatic traversal of neighboring problems around a focal problem, surfacing novel inspirations to users. In a study of the efficacy of the methods, over 60\% of ``inspirations'' retrieved this way were found to be useful and novel---a relative boost of 50-60\% over the best-performing baselines. For example, given an input problem of reminding patients to take medication, our system retrieves related problems such as in patient health tracking and alerting devices.  

\begin{figure}
\includegraphics[width=\columnwidth]{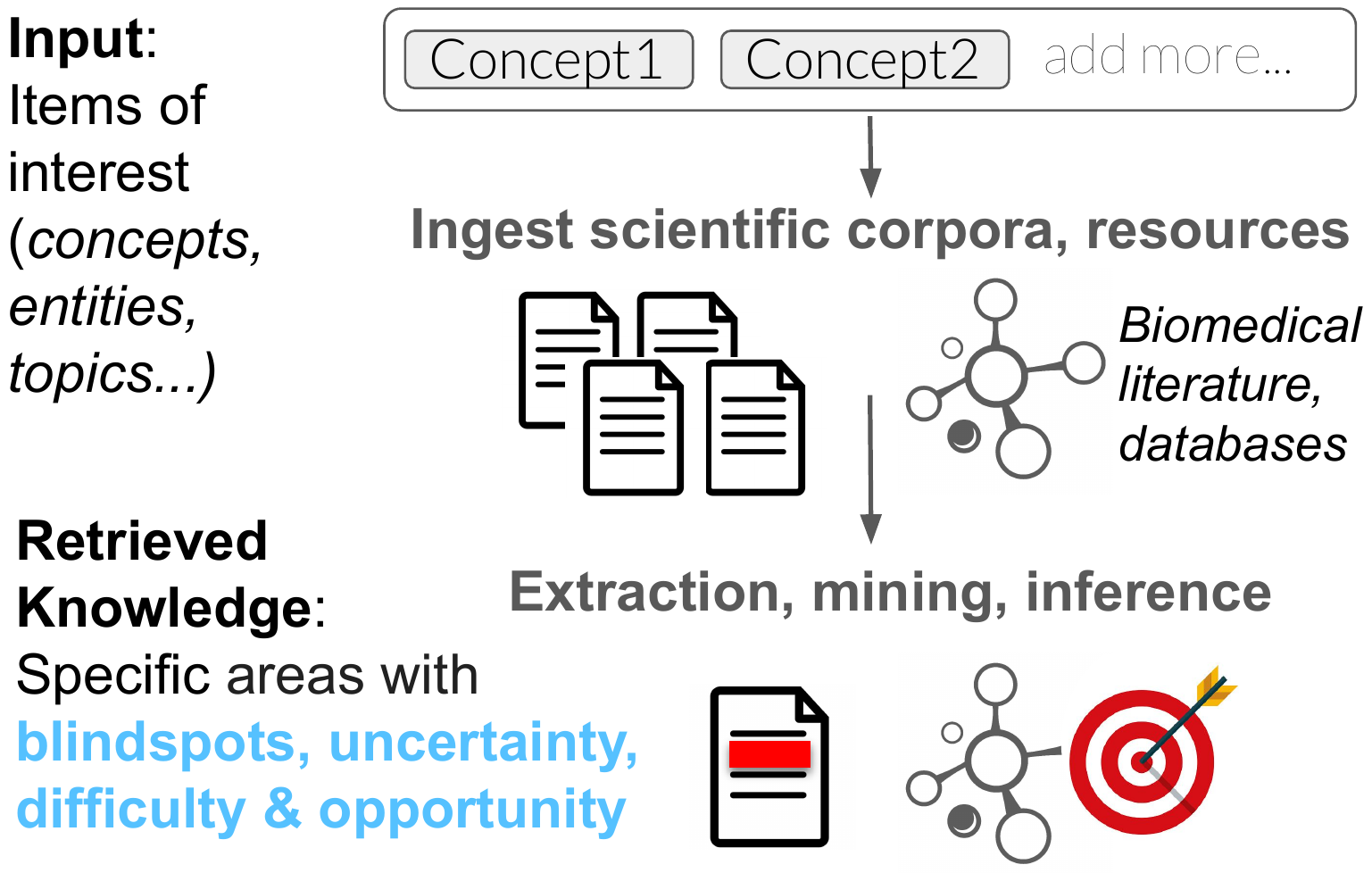}
    \caption{Suggesting research opportunities for query concepts (e.g., medical topics) by identifying blindspots, gaps in collective knowledge and promising areas for exploration.}
    \label{fig:attention}
\end{figure}

\textbf{Guiding attention and problem identification.} We see great opportunity in developing methods for guiding the attention of researchers to important areas in the space of ideas where there exists less knowledge or certainty \cite{lahav2021search,AttentionBiases} (Figure \ref{fig:attention}). 
In one direction, we built a search engine that allows users to retrieve {outer knowledge} in the form of difficulties, uncertainties and hypotheses in the literature. The key goals of this mode of search are to bolster attention to rising and standing challenges of relevance to the user so as to help overall with identification and selection of problems. We performed experiments with participants with diverse research backgrounds, including medical doctors working in a large hospital. Using query topics as a proxy for the inner world of participants' interests, we found the system could dramatically outperform PubMed search, the go-to biomedical search engine, at discovering important and interesting areas of challenges and directions. For example, while searching PubMed for the ACE2 receptor in the context of COVID-19 returns well-studied results, the prototype system by contrast focuses on finding statements of uncertainty, open questions and initial hypotheses, like a paper noting the {\em possibility} that ACE2 plays a role in liver damage in COVID-19 patients.

Another direction on biases and blindspots considers the long-term effort to identify protein-protein interactions (PPIs). 
A dataset of the growing graph of confirmed PPIs over decades was constructed and leveraged to identify patterns of scientific attention \cite{AttentionBiases}.  A temporal analysis revealed a significant ``bias of locality,'' where explorations of PPIs are launched more frequently from those that were most recently studied, rather than following more general prioritization of exploration. While locality reflects an understandable focus on adjacent and connected problems in the biosciences, the pattern of attention leads to systematic blindspots in large, widely used PPI databases that are likely unappreciated---further exacerbating attentional biases.  The study further demonstrated mechanisms for reprioritizing candidate PPIs based on properties of proteins, and showed how earlier discoveries could be made with use of the debiasing methods. The findings underscore the promise of tools that retrieve existing outer world knowledge to guide attention to worthwhile directions. In this case, the outer knowledge source is a PPI \emph{database}, and a user-selected sub-graph provides a proxy for inner world knowledge and interests.

\textbf{Literature search and review}.
A great body of work on \emph{literature search and review} has deep relevance to task-guided retrieval in the sciences. In particular, we see great opportunity with building on recent advances in information retrieval to (1) help biomedical researchers with domain-specific representations and (2) enhance scientific search by building new neural models. 

Specialized search systems have been developed for the biomedical domain, with the overall vision of harnessing natural language understanding technologies to help researchers discover relevant evidence and expedite the costly process of systematic literature review \cite{nye2021understanding}. For example, \citet{nye2021understanding} build a search and synthesis system based on automated extraction of biomedical treatment-outcome relations from clinical trial reports. The system is found to assist in identification of drug repurposing opportunities.  As another recent example, the SPIKE system enables researchers to extract and retrieve facts from a corpus using an expressive query language with biomedical entity types and new term classes that the user can define interactively  \cite{taub-tabib-etal-2020-interactive}.  Together, this work underscores the importance of extracting a semantically meaningful representation of outer world knowledge that aligns with core aspects of inner world reasoning by researchers (see Section \ref{sec:challenges}).

In separate work, neural language models built via self-supervision on large corpora of biomedical publications have recently led to performance boosts and new features in literature search systems  \cite{wang2021domain}, such as support for natural language queries that provide users with a more natural way to formulate their informational goals. Neural models have also been trained to match abstract discourse aspects of pairs of papers (e.g., sentences referring to methodologies) and automatically retrieve documents that are aspectually similar \cite{mysore2022multi}. By employing a representation that aligns with scientific reasoning across areas, this method achieves state-of-art results across biomedical and computer science literature.

\begin{figure}
    \centering
    \includegraphics[width=0.95\columnwidth]{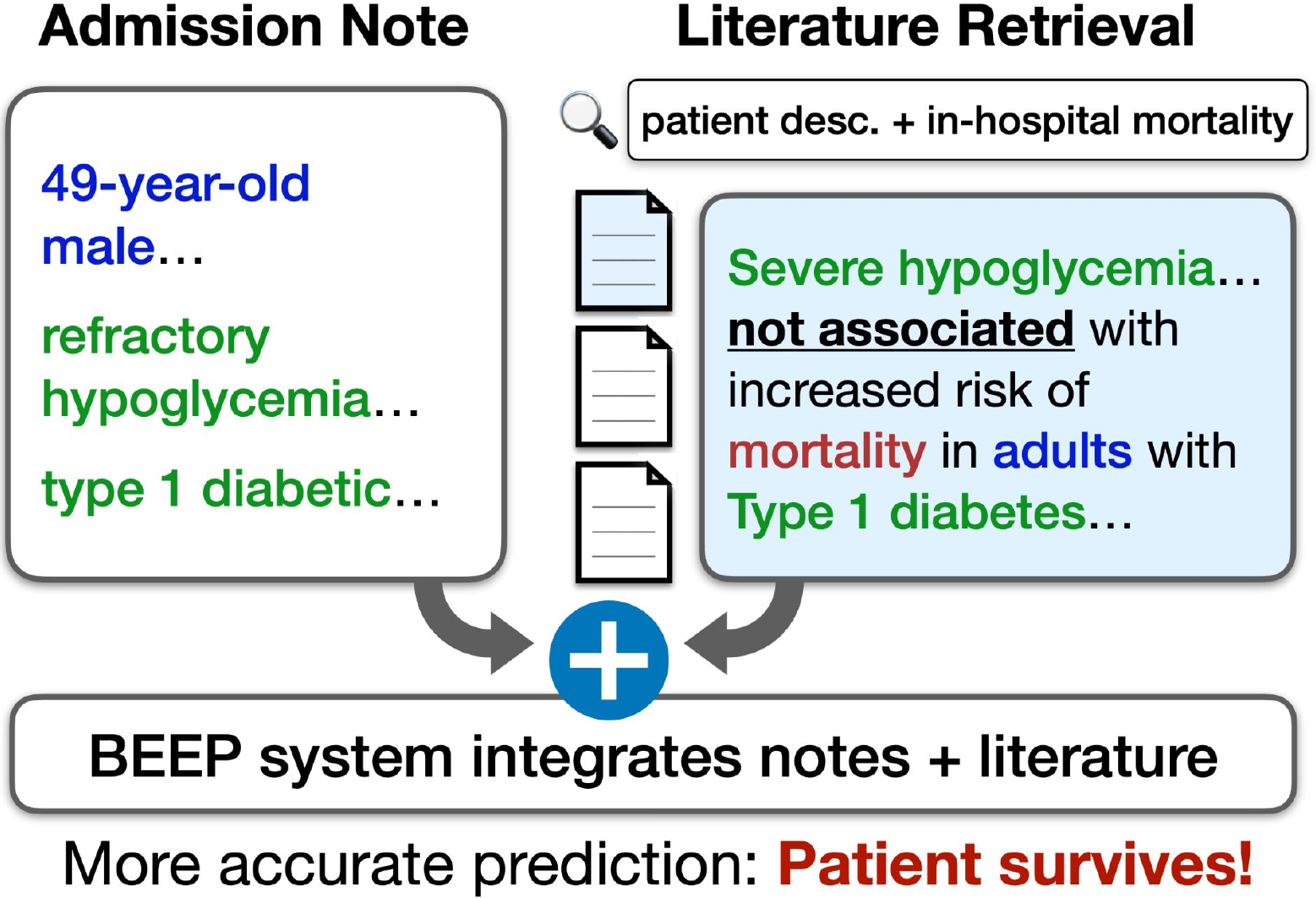}
    \caption{Leveraging medical corpora to enhance the precision of AI models for inference about patient outcomes.}
    \label{fig:prediction}
\end{figure}

\textbf{Experimentation, analysis, and action}. Beyond helping researchers via awareness and knowledge, we see great opportunities to use scientific corpora to construct task-centric inferential systems with automated models and tools for assisting with analysis, prediction and decisions. We demonstrate these ideas by casting two different lines of work as cases of task-guided retrieval. 

(1) \emph{Workflows} are multi-step computational pipelines used as part of scientific experimentation for data preparation, analysis and simulation \cite{gil2022will}. Technically this includes execution of code scripts, services and tools, querying databases and submitting jobs to the cloud. In the life sciences, in areas such as genomics, there are specialized workflow management systems to help researchers find and use workflows, enabled by a community that creates and publicly shares repositories of workflows with standardised interfaces, metadata and functional annotations of tools and data. As discussed in \citet{gil2022will}, machine learning algorithms can potentially use these resources to automate workflow construction, learning to retrieve and synthesize data analysis pipelines. In this setting, outer world knowledge takes the form of workflow repositories, from which systems retrieve and synthesize modular building blocks; users' inner world is reflected via analysis objectives and constraints.

(2) In our work on clinical predictions \cite{naik2021literature},   
the goal is to enhance prediction of medical outcomes of patients hospitalized in the intensive care unit (ICU), such as in-hospital mortality or prolonged length of stay. Our system, named BEEP (biomedical evidence enhanced prediction), learns to perform predictions by retrieving medical papers that are relevant to each specific ICU patient, and to synthesize this outer knowledge in combination with internal EMR knowledge to form a final prediction. 
The primary envisaged user is a practice-oriented researcher---a medical doctor, whose inner knowledge is given by a rough proxy in the form of internal clinical notes from which we extract ``queries'' issued over medical papers. We find BEEP to provide large improvements over state-of-art models that do not use retrieval from the literature. BEEP's output can be aligned with inner world representations, e.g., matches between patient aspects and related cohorts in papers (see Figure \ref{fig:prediction}).

\textbf{Learning and understanding}. We introduced a system \cite{accord2022} for helping users learn about \emph{new} concepts by showing definitions grounded in \emph{familiar} concepts; e.g., a new algorithm is explained as a variant of an algorithm familiar to the user. Cognitive studies have asserted that effective descriptions of a new concept ground it within the network of known concepts. Our system takes as input a list of \emph{source concepts} reflecting the user's inner knowledge as obtained from papers that they have written or read. When the user seeks a definition of a new target concept, we retrieve outer knowledge in the form of definitions appearing in scientific papers in which the \emph{target concept} is explained in terms of the {source concepts}; a neural text generation model then re-writes the text in a structured, template form that relates the target to the source.

\section{Opportunities Ahead}
\label{sec:challenges}

The challenges of task-guided retrieval in support of researchers frame a host of problems and opportunities. We discuss select challenges and directions (see also Table \ref{tab:challenges}). We begin with an illustrative example, imagining a futuristic system to motivate the discussion.

\subsection{Aspirations}

We envision tools that flow outer world knowledge to researchers based on inferences about their inner world---users' knowledge, past and present goals and difficulties, and the tasks from Table \ref{tab:tasks} they are engaged in. The systems would use multiple signals for making inferences, including users' papers, data, experiments and communication channels, and also converse with the user to understand needs and suggest solutions, hypotheses and experiments.

We foresee systems powered by rich representations of both inner and outer scientific knowledge. For a given concept, e.g., a certain algorithm or organism, an aspirational system would ingest all papers on the subject to form a multi-faceted representation of concepts as objects with associated properties and functions. Using this representation, the system could assist in \emph{literature search and review}, enabling expressive queries to outer world information that target abstract aspects like functionalities, mechanisms, behaviors and designs in a manner that transcends field-specific jargon, abstracting away lexical differences that hindered historical search engines (e.g., Google Scholar). To help users \emph{learn and understand new concepts} they encounter, the system would explain them in relation to other concepts the user already knows. A future system might also assist in automating \emph{experimentation, analysis and action} and in \emph{forming directions}, by forming compositions of concepts and predicting the resultant affordances; for example, matching a certain algorithm with a suitable problem based on the algorithm’s properties and the problem’s requirements, matching an organism with a specific method of measurement or modification, or recombining parts of two devices to form a new device. The system could help identify related problems in the literature, synthesizing from them useful suggestions for problem reformulations. 
Considering the huge combinatorial space of potential suggestions, a system could assist in \emph{prioritization} using estimated measures of interestingness, feasibility and value by synthesizing historical and current signals in literature, online discussions and knowledge bases. 

Envisioned systems would be designed as human-centric, focusing on the individual researcher. The systems would enable users to convey preferences, goals and interests, and mediate the presentation of suggested directions and problem solutions based on personal prior knowledge, proposing concrete new directions grounded in representations that researchers can follow, and assisting users in \emph{reading} complex retrieved texts by editing their language to conform with concepts that users are familiar with.  

\vspace{-.2in}

\subsection{Research Directions}

\begin{table*}
{
\begin{tabular}{p{0.3\linewidth}|p{0.66\linewidth}}
    \hline
    \textbf{Challenge} & \textbf{Description}
    \\
    \hline
    \parbox[t]{\linewidth}{\emph{Task-aligned representations, scientific NLP}}
    &\parbox[t]{\linewidth}{How can we automatically and accurately extract conceptual representations of scientific knowledge, that are aligned with tasks that comprise the endeavor of science (Table \ref{tab:tasks})? How can we build NLP models that understand full scientific papers?}
    \\
    \hline 
    
        \parbox[t]{\linewidth}{\emph{Computational algebra of ideas}}
    &\parbox[t]{\linewidth}{Can we build representations of scientific knowledge that support composition of ideas? e.g., inferring the result of recombining two concepts.}
        \\
    \hline
    
     \parbox[t]{\linewidth}{\emph{Identifying conceptual relationships across literature}}
    &\parbox[t]{\linewidth}{How do we detect important relationships across scientific ideas, across related discussions in different communities and areas? How can we resolve challenges of diversity, ambiguity, and multiple levels of detail in scientific language?} 
    \\
    \hline
        \parbox[t]{\linewidth}{\emph{Estimation of personal knowledge}}
    &\parbox[t]{\linewidth}{How can we estimate the knowledge of a given researcher? What are useful, practical models of this knowledge? What concepts does a researcher know, which of their aspects, and to what technical extent? How do we account for \emph{latent} knowledge? 
}
        \\
        \hline
         \parbox[t]{\linewidth}{\emph{Addressing gaps in knowledge}}
    &\parbox[t]{\linewidth}{Given an estimated model of a researcher’s knowledge, and given a specific task in Table \ref{tab:tasks}, what new knowledge would be useful for the task at hand?}
        \\
        \hline 
        
         \parbox[t]{\linewidth}{\emph{Estimation of preferences, goals, interests}}
    &\parbox[t]{\linewidth}{How can we estimate key latent preferences, interests and subjective utilities of researchers? Using information in papers and discussions to infer factors behind researchers' choices.}
        \\
        \hline

    \parbox[t]{\linewidth}{\emph{Prediction and prioritization }}
    &\parbox[t]{\linewidth}{How might we identify promising sparse/unexplored areas in large “spaces of ideas” and prioritize directions that are novel, plausible and valuable?
}
        \\
        \hline
    \parbox[t]{\linewidth}{\emph{Developing new representations for learning and communicating}}
    &\parbox[t]{\linewidth}{Might the way we read and write papers change to be more effective? Might we communicate with machine-actionable, interlinked representations of scholarly knowledge. Might we create personalized ``living'' documents that tailor their content to readers' backgrounds.}
    \\
    \hline
\end{tabular}
}
\caption{Directions with formulating and leveraging computational representations of scientific knowledge.}
\label{tab:challenges}
\end{table*}

While we have witnessed remarkable strides in AI, the journey towards actualizing our vision requires further advancement. Envisioning such capabilities, however, can serve as a compass for directing research endeavors. An encouraging development can be seen in the recent developments with large language models, which have demonstrated surprising capabilities with interpreting and generating complex texts and tackling technical tasks. The proficiencies demonstrated by these models instills confidence that many of the possibilities that we discussed are attainable.  We now elaborate on challenges and directions ahead, including limitations in representing scientific knowledge and making inferences about the inner worlds of researchers (see Table \ref{tab:challenges}).

\textbf{Task-aligned representations and scientific NLP.} Paul Thagard writes: ``thinking can best be understood in terms of representational structures in the mind and computational procedures that operate on those structures''.  We seek representations that can be aligned with human thinking---for insight-building, decision making and communication. Can we go beyond textual representation toward representations that support such cognitive processes?

The quest for a \emph{universal schema} representing scientific ideas goes back hundreds of years. Gottfried Leibniz and René Descartes were intrigued by the prospects of a universal codification of knowledge. Leibniz proposed the \emph{characteristica universalis}, a hypothesized formal language of ideas enabling inferences with algebraic operators. While such a representation is not within reach, envisioning its existence---and how to even roughly approximate it---points to important research directions. One exciting direction is obtaining representations that support a ``computational algebra of ideas''---e.g., modeling compositions of concepts and the affordances that would be formed as a result. Early work on learning vector representations of natural language concepts supported rudimentary forms of addition, subtraction, and analogy (e.g., the Word2vec model). 

Recently, large language models (LLM) \cite{openai2023gpt4} have made striking progress in generating new content and coherently combining concepts. Emerging evidence on GPT-4's ability to reason not only in unstructured language but also with logical structures grounded in code, suggests strong potential for generating novel ideas via compositionality and relational reasoning \cite{bubeck2023sparks}. Our early experiments with GPT-4 have revealed a constellation of promising abilities to assist with the scientific process, such as formulating hypotheses, recommending future research directions, and critiquing studies. Equipped with training and retrieval with access to millions of scientific papers, descendants of today's models may have an ability to synthesize original scientific concepts with the in-depth technical detail at a level reported in high-quality scientific papers. We see great opportunity ahead to leverage LLMs to augment human scientific reasoning along the lines described in this paper. 

One limitation with LLMs is that representations learned by these models are currently far from understood and lack ``hooks'' for control and interpretability, which are important in human-AI collaboration. In line with our focus on grounding representations of outer world knowledge with inner world cognitive aspects, we have pursued methods that ``reverse engineer'' scientific papers to automatically extract, using NLP, \emph{structured } representations that balance three desiderata: 

\begin{enumerate}[wide, labelindent=0pt]
\item Semantically meaningful representations, \emph{aligned with a salient task} from the tasks in Table \ref{tab:tasks}, grounded in cognitive research to guide us toward useful structures. 
\item Representations with \emph{sufficient level of abstraction} to generalize across areas and topics.
\item Representations \emph{expressive enough for direct utility} in helping researchers as measured in human studies.
\end{enumerate} 

For example, we have extracted representations of \emph{causal mechanisms} and \emph{hierarchical graphs of functional relationships}. This kind of decomposition of ideas has enabled us to perform basic analogical inference in the space of technological and scientific ideas, helping researchers discover inspirations (see Section \ref{sec:taskguided}). 
However, many richer structures should be explored (e.g., of experimentation processes and methodologies, to enable tasks in Table \ref{tab:tasks}). 

A central challenge is that current models' extraction accuracy is limited, and the {diversity of scientific language} leads to problems in generalization and {normalization of terms and concepts}. 
We have pursued construction of new datasets, models and evaluations for identifying similarity between concepts and aspects across papers \cite{cattan2021scico,mysore2022multi}, with fundamental problems in resolving \emph{diversity, ambiguity and hierarchy} of language. As our results have highlighted, models tend to focus on surface-level lexical patterns, rather than deeper semantic relationships. 
Generally, substantial advances are needed to handle challenges posed by scientific documents. We require NLP models with full-document understanding, not only of text but of tables, equations, figures, and reference links. Open access corpora (e.g., S2ORC \cite{lo-wang-2020-s2orc}) provide a foundation to address this challenge.

\textbf{New modes of writing and reading.} Perhaps {the way we write} can be dramatically different, using machine-actionable representations? Beyond reporting and documentation, writing represents a channel between the inner and outer worlds, forcing us to communicate ideas in concrete language; this process often begets new questions and perspectives. Can systems accompany different phases of writing, suggesting new ideas? In parallel, there is the task of \emph{reading} what others have written; a recent interactive PDF reader offers, for example, customized concept definitions \cite{reader2021}. We imagine a future where every reader will see a different form of the same paper, re-written to align with readers' knowledge; e.g., 
our personalized concept definitions system \cite{accord2022} (§\ref{sec:taskguided}) will insert new wording and explanations grounded in readers' knowledge. 

\textbf{Internal world of researchers.} Grounding new concepts in readers' knowledge, suggests a wider and highly challenging problem. How can we enable researchers to specify their knowledge and preferences to direct systems to carry out tasks? Directly querying for these aspects burdens the researcher and may be prone to reporting biases. Digital traces present an opportunity for {\em automatically} estimating a researcher's knowledge, objectives, needs and interests---based on data. We are interested in using researchers' papers to estimate what concepts users know and to what extent. We envision mixed-initiative interfaces \cite{horvitz1999principles} in which approximations of the inner world are presented to researchers and refined in human-machine collaboration, to identify and fill personal gaps in knowledge for a specific task. Representations of interest and preference are central in web commerce based on user activity histories. We are encouraged by results highlighting the feasibility of rich user models, e.g., in search personalization \cite{teevan2005personalizing, shah2022situating} and dynamic inferences \cite{horvitz2013lumiere}. Paul Samuelson wrote of ``revealed preferences''--- preferences revealed indirectly by the economic price people are willing to pay; while not equivalent, researchers' digital traces may reveal preferences, e.g., by working on one problem and not another. 

\textbf{Prediction and prioritization of directions.} Whenever we decide to work on a research direction, we are implicitly making a prediction about an area in ``idea space''. Can automated systems help make these predictions? This involves identifying promising areas and generating directions---hypotheses, ideas---in either natural or structured language, under constraints on users' background knowledge; directions should be ranked by estimated likelihood (feasibility, plausibility), utility and novelty. Despite the great challenges involved, we are encouraged by advances in models trained for predicting specific targets (e.g., protein structures \cite{jumper2021highly}); we see potential in building on these advances as part of our \emph{wider} agenda that considers the inner world of cognitive aspects and tasks, and the outer world outside the context of a narrow dataset.

\textbf{Pursuing challenges of translation}. Finally, we note challenges for introducing new technologies into scientific workflows. In the context of systems for discovery, researchers interviewed in our studies \cite{portenoy2021bridger} reported being limited in time and resources, making them less likely to enter new areas and learn unfamiliar concepts, preventing them discovering potentially promising ideas. More broadly, the sociotechnical environment in which AI models are deployed has critical impact on their success \cite{horvitz21future,lohr2021ever}. A pertinent example comes via reports on difficulties with translating IBM's Watson Health systems into practice. The vision of the effort included systems providing insights about patients by mining research papers to suggest, e.g., therapies or diagnostics \cite{lohr2021ever}. A prototype system faced difficulties ranging from data processing and availability problems to deeper perceived gaps between the system's understanding of literature and that of physicians \cite{thamba2022watson}. Challenges such as these are fundamental to the fielding of new applications not only in healthcare but in any setting where humans are required to interact with AI systems \cite{weld2019challenge}. While  issues such as data quality and privacy are orthogonal to our agenda, we see directions in modeling of human needs and limitations to inform the design of human-AI experiences within scientific workflows. 

\section{Summary}
As the terrain of science widens at a fast pace, researchers are constrained by the limits of human cognition, and lack principled methods to follow developments, guide attention, and formulate and prioritize directions. 
For the first time in history, essentially all of scientific knowledge and discourse has moved into the digital space. At the time of this writing, dramatic advances in AI with large language models are taking place at breathtaking speed. These shifts present tremendous opportunities for leveraging scientific corpora as databases from which solutions, insights, and inspirations can be gleaned. We see opportunity ahead for systems that can address the imbalance between the treasure trove of scholarly information and researchers' limited ability to reach into it, harnessing humankind's collective knowledge to revolutionize the scientific process. Numerous challenges stand in the way of the vision we have laid out.  However, even small steps forward will unlock vast opportunities for making advances at the frontiers of science.

\begin{acks}
We thank the members of the Semantic Scholar team for stimulating discussions. Projects were supported by NSF Convergence Accelerator Grant 2132318, NSF RAPID grant 2040196, and ONR grant {N00014-18-1-2193}.
\end{acks}

\bibliographystyle{plainnat}
\bibliography{sample-base}

\begin{thebibliography}{40}
\providecommand{\natexlab}[1]{#1}
\providecommand{\url}[1]{\texttt{#1}}
\expandafter\ifx\csname urlstyle\endcsname\relax
  \providecommand{\doi}[1]{doi: #1}\else
  \providecommand{\doi}{doi: \begingroup \urlstyle{rm}\Url}\fi

\bibitem[Bubeck et~al.(2023)Bubeck, Chandrasekaran, Eldan, Gehrke, Horvitz,
  Kamar, Lee, Lee, Li, Lundberg, et~al.]{bubeck2023sparks}
S{\'e}bastien Bubeck, Varun Chandrasekaran, Ronen Eldan, Johannes Gehrke, Eric
  Horvitz, Ece Kamar, Peter Lee, Yin~Tat Lee, Yuanzhi Li, Scott Lundberg,
  et~al.
\newblock Sparks of artificial general intelligence: Early experiments with
  {GPT}-4.
\newblock \emph{arXiv preprint arXiv:2303.12712}, 2023.

\bibitem[Cattan et~al.(2021)Cattan, Johnson, Weld, Dagan, Beltagy, Downey, and
  Hope]{cattan2021scico}
Arie Cattan, Sophie Johnson, Daniel Weld, Ido Dagan, Iz~Beltagy, Doug Downey,
  and Tom Hope.
\newblock Scico: Hierarchical cross-document coreference for scientific
  concepts.
\newblock \emph{Automated Knowledge Base Construction (AKBC) 2021}, 2021.

\bibitem[Chan et~al.(2018)Chan, Chang, Hope, Shahaf, and
  Kittur]{chan2018solvent}
Joel Chan, Joseph~Chee Chang, Tom Hope, Dafna Shahaf, and Aniket Kittur.
\newblock Solvent: A mixed initiative system for finding analogies between
  research papers.
\newblock \emph{Proceedings of the ACM on Human-Computer Interaction},
  2\penalty0 (CSCW):\penalty0 1--21, 2018.

\bibitem[Chu and Evans(2021)]{chu2021slowed}
Johan~SG Chu and James~A Evans.
\newblock Slowed canonical progress in large fields of science.
\newblock \emph{Proceedings of the National Academy of Sciences}, 118\penalty0
  (41), 2021.

\bibitem[Garc{\'\i}a-Villar(2021)]{garcia2021critical}
Cristina Garc{\'\i}a-Villar.
\newblock A critical review on altmetrics: can we measure the social impact
  factor?
\newblock \emph{Insights into Imaging}, 12\penalty0 (1):\penalty0 1--10, 2021.

\bibitem[Gil(2022)]{gil2022will}
Yolanda Gil.
\newblock Will {AI} write scientific papers in the future?
\newblock \emph{AI Magazine}, 2022.

\bibitem[Head et~al.(2021)Head, Lo, Kang, Fok, Skjonsberg, Weld, and
  Hearst]{reader2021}
Andrew Head, Kyle Lo, Dongyeop Kang, Raymond Fok, Sam Skjonsberg, Daniel~S.
  Weld, and Marti~A. Hearst.
\newblock Augmenting scientific papers with just-in-time, position-sensitive
  definitions of terms and symbols.
\newblock In \emph{Proceedings of the 2021 CHI Conference on Human Factors in
  Computing Systems}, 2021.

\bibitem[Hope et~al.(2017)Hope, Chan, Kittur, and Shahaf]{hope2017accelerating}
Tom Hope, Joel Chan, Aniket Kittur, and Dafna Shahaf.
\newblock Accelerating innovation through analogy mining.
\newblock In \emph{Proceedings of the 23rd ACM SIGKDD International Conference
  on Knowledge Discovery and Data Mining}. ACM, 2017.

\bibitem[Hope et~al.(2020)Hope, Portenoy, Vasan, Borchardt, Horvitz, Weld,
  Hearst, and West]{hope2020scisight}
Tom Hope, Jason Portenoy, Kishore Vasan, Jonathan Borchardt, Eric Horvitz,
  Daniel~S Weld, Marti~A Hearst, and Jevin West.
\newblock Scisight: Combining faceted navigation and research group detection
  for covid-19 exploratory scientific search.
\newblock In \emph{EMNLP}, 2020.

\bibitem[Hope et~al.(2021)Hope, Amini, Wadden, van Zuylen, Parasa, Horvitz,
  Weld, Schwartz, and Hajishirzi]{hope2021mechanisms}
Tom Hope, Aida Amini, David Wadden, Madeleine van Zuylen, Sravanthi Parasa,
  Eric Horvitz, Daniel~S Weld, Roy Schwartz, and Hannaneh Hajishirzi.
\newblock Extracting a knowledge base of mechanisms from covid-19 papers.
\newblock In \emph{NAACL}, 2021.

\bibitem[Hope et~al.(2022)Hope, Tamari, Kang, Hershcovich, Chan, Kittur, and
  Shahaf]{hope2021scaling}
Tom Hope, Ronen Tamari, Hyeonsu Kang, Daniel Hershcovich, Joel Chan, Aniket
  Kittur, and Dafna Shahaf.
\newblock Scaling creative inspiration with fine-grained functional facets of
  product ideas.
\newblock In \emph{CHI}, 2022.

\bibitem[Horvitz(1999)]{horvitz1999principles}
Eric Horvitz.
\newblock Principles of mixed-initiative user interfaces.
\newblock In \emph{Proceedings of the SIGCHI conference on Human Factors in
  Computing Systems}, pages 159--166, 1999.

\bibitem[Horvitz(2021)]{horvitz21future}
Eric Horvitz.
\newblock The future of biomedical informatics: Bottlenecks and opportunities.
\newblock In \emph{Biomedical Informatics: Computer Applications in Health Care
  and Biomedicine, E.H. Shortliffe, J.J. Cimino, et. al}. Springer, 2021.

\bibitem[Horvitz et~al.(1998)Horvitz, Breese, Heckerman, Hovel, and
  Rommelse]{horvitz2013lumiere}
Eric~J Horvitz, John~S Breese, David Heckerman, David Hovel, and Koos Rommelse.
\newblock The {L}umiere project: Bayesian user modeling for inferring the goals
  and needs of software users.
\newblock In \emph{Proceedings of the Conference on Uncertainty in AI}, pages
  256--265, 1998.

\bibitem[Jumper et~al.(2021)Jumper, Evans, Pritzel, Green, Figurnov,
  Ronneberger, Tunyasuvunakool, Bates, {\v{Z}}{\'\i}dek, Potapenko,
  et~al.]{jumper2021highly}
John Jumper, Richard Evans, Alexander Pritzel, Tim Green, Michael Figurnov,
  Olaf Ronneberger, Kathryn Tunyasuvunakool, Russ Bates, Augustin
  {\v{Z}}{\'\i}dek, Anna Potapenko, et~al.
\newblock Highly accurate protein structure prediction with alphafold.
\newblock \emph{Nature}, 596\penalty0 (7873):\penalty0 583--589, 2021.

\bibitem[King et~al.(2004)King, Whelan, Jones, Reiser, Bryant, Muggleton, Kell,
  and Oliver]{king2004functional}
Ross~D King, Kenneth~E Whelan, Ffion~M Jones, Philip~GK Reiser, Christopher~H
  Bryant, Stephen~H Muggleton, Douglas~B Kell, and Stephen~G Oliver.
\newblock Functional genomic hypothesis generation and experimentation by a
  robot scientist.
\newblock \emph{Nature}, 427\penalty0 (6971):\penalty0 247--252, 2004.

\bibitem[Kuhn(1970)]{kuhn1970structure}
Thomas~S Kuhn.
\newblock \emph{The structure of scientific revolutions}, volume 111.
\newblock Chicago University of Chicago Press, 1970.

\bibitem[Lahav et~al.(2022)Lahav, Falcon, Kuehl, Johnson, Parasa, Shomron,
  Chau, Yang, Horvitz, Weld, and Hope]{lahav2021search}
D~Lahav, JS~Falcon, B~Kuehl, S~Johnson, S~Parasa, N~Shomron, DH~Chau, D~Yang,
  E~Horvitz, DS~Weld, and T~Hope.
\newblock A search engine for discovery of scientific challenges and
  directions.
\newblock In \emph{AAAI}, 2022.

\bibitem[Langley et~al.(1987)Langley, Simon, Bradshaw, and
  Zytkow]{langley1987scientific}
Pat Langley, Herbert~A Simon, Gary~L Bradshaw, and Jan~M Zytkow.
\newblock \emph{Scientific discovery: Computational explorations of the
  creative processes}.
\newblock MIT press, 1987.

\bibitem[Lo et~al.(2020)Lo, Wang, Neumann, Kinney, and
  Weld]{lo-wang-2020-s2orc}
Kyle Lo, Lucy~Lu Wang, Mark Neumann, Rodney Kinney, and Daniel~S. Weld.
\newblock {S2ORC: The Semantic Scholar Open Research Corpus}.
\newblock In \emph{Proceedings of ACL}, 2020.

\bibitem[Lohr(2021)]{lohr2021ever}
Steve Lohr.
\newblock What ever happened to ibm’s watson.
\newblock \emph{The New York Times}, 16\penalty0 (7):\penalty0 21, 2021.

\bibitem[Murthy et~al.(2022)Murthy, Lo, Bhagavatula, King, Kuehl, Johnson,
  Borchardt, Weld, Hope, and Downey]{accord2022}
Sonia Murthy, Kyle Lo, Chandra Bhagavatula, Daniel King, Bailey Kuehl, Sophie
  Johnson, Jonathan Borchardt, Daniel~S. Weld, Tom Hope, and Doug Downey.
\newblock Accord: A multi-document approach to generating diverse descriptions
  of scientific concepts.
\newblock In \emph{EMNLP}, 2022.

\bibitem[Mysore et~al.(2022)Mysore, Cohan, and Hope]{mysore2022multi}
Sheshera Mysore, Arman Cohan, and Tom Hope.
\newblock Multi-vector models with textual guidance for fine-grained scientific
  document similarity.
\newblock \emph{NAACL}, 2022.

\bibitem[Naik et~al.(2022)Naik, Parasa, Feldman, Wang, and
  Hope]{naik2021literature}
Aakanksha Naik, Sravanthi Parasa, Sergey Feldman, Lucy~Lu Wang, and Tom Hope.
\newblock Literature-augmented clinical outcome prediction.
\newblock \emph{NAACL}, 2022.

\bibitem[Nori et~al.(2023)Nori, King, McKinney, Carignan, and
  Horvitz]{nori2023capabilities}
Harsha Nori, Nicholas King, Scott~Mayer McKinney, Dean Carignan, and Eric
  Horvitz.
\newblock Capabilities of {GPT}-4 on medical challenge problems.
\newblock \emph{arXiv preprint arXiv:2303.13375}, 2023.

\bibitem[Nuzzo et~al.(2015)]{nuzzo2015scientists}
Regina Nuzzo et~al.
\newblock How scientists fool themselves--and how they can stop.
\newblock \emph{Nature}, 526\penalty0 (7572):\penalty0 182--185, 2015.

\bibitem[Nye et~al.(2021)Nye, DeYoung, Lehman, Nenkova, Marshall, and
  Wallace]{nye2021understanding}
Benjamin Nye, Jay DeYoung, Eric Lehman, Ani Nenkova, Iain~J Marshall, and
  Byron~C Wallace.
\newblock Understanding clinical trial reports: Extracting medical entities and
  their relations.
\newblock In \emph{AMIA Annual Symposium Proceedings}, volume 2021, page 485.
  American Medical Informatics Association, 2021.

\bibitem[OpenAI(2023)]{openai2023gpt4}
OpenAI.
\newblock Gpt-4 technical report, 2023.

\bibitem[Portenoy et~al.(2022)Portenoy, Radensky, West, Horvitz, Weld, and
  Hope]{portenoy2021bridger}
Jason Portenoy, Marissa Radensky, Jevin West, Eric Horvitz, Daniel Weld, and
  Tom Hope.
\newblock Bridger: Toward bursting scientific filter bubbles and boosting
  innovation via novel author discovery.
\newblock \emph{CHI}, 2022.

\bibitem[Pyzer-Knapp et~al.(2022)Pyzer-Knapp, Pitera, Staar, Takeda, Laino,
  Sanders, Sexton, Smith, and Curioni]{pyzer2022accelerating}
Edward~O Pyzer-Knapp, Jed~W Pitera, Peter~WJ Staar, Seiji Takeda, Teodoro
  Laino, Daniel~P Sanders, James Sexton, John~R Smith, and Alessandro Curioni.
\newblock Accelerating materials discovery using artificial intelligence, high
  performance computing and robotics.
\newblock \emph{npj Computational Materials}, 8\penalty0 (1):\penalty0 1--9,
  2022.

\bibitem[Shah and Bender(2022)]{shah2022situating}
Chirag Shah and Emily~M Bender.
\newblock Situating search.
\newblock In \emph{ACM SIGIR Conference on Human Information Interaction and
  Retrieval}, pages 221--232, 2022.

\bibitem[Singer et~al.(2020)Singer, Radinsky, and Horvitz]{AttentionBiases}
Uriel Singer, Kira Radinsky, and Eric Horvitz.
\newblock {On biases of attention in scientific discovery}.
\newblock \emph{Bioinformatics}, 12 2020.
\newblock URL \url{https://doi.org/10.1093/bioinformatics/btaa1036}.

\bibitem[Swanson(1986)]{Swanson1986-SWAFOR}
D.~R. Swanson.
\newblock Fish oil, raynaud's syndrome, and undiscovered public knowledge.
\newblock \emph{Perspectives in Biology and Medicine}, 30\penalty0
  (1):\penalty0 7--18, 1986.

\bibitem[Taub~Tabib et~al.(2020)Taub~Tabib, Shlain, Sadde, Lahav, Eyal, Cohen,
  and Goldberg]{taub-tabib-etal-2020-interactive}
Hillel Taub~Tabib, Micah Shlain, Shoval Sadde, Dan Lahav, Matan Eyal, Yaara
  Cohen, and Yoav Goldberg.
\newblock Interactive extractive search over biomedical corpora.
\newblock In \emph{Proceedings of the 19th SIGBioMed Workshop on Biomedical
  Language Processing}, pages 28--37, Online, July 2020. Association for
  Computational Linguistics.
\newblock \doi{10.18653/v1/2020.bionlp-1.3}.
\newblock URL \url{https://aclanthology.org/2020.bionlp-1.3}.

\bibitem[Teevan et~al.(2005)Teevan, Dumais, and
  Horvitz]{teevan2005personalizing}
Jaime Teevan, Susan~T Dumais, and Eric Horvitz.
\newblock Personalizing search via automated analysis of interests and
  activities.
\newblock In \emph{Proceedings of the 28th annual international ACM SIGIR
  conference on Research and development in information retrieval}, pages
  449--456, 2005.

\bibitem[Thagard(2012)]{thagard2012cognitive}
Paul Thagard.
\newblock \emph{The cognitive science of science: Explanation, discovery, and
  conceptual change}.
\newblock Mit Press, 2012.

\bibitem[Thamba and Gunderman(2022)]{thamba2022watson}
Aish Thamba and Richard~B Gunderman.
\newblock For watson, solving cancer wasn't so elementary: Prospects for
  artificial intelligence in radiology.
\newblock \emph{Academic Radiology}, 29\penalty0 (2):\penalty0 312--314, 2022.

\bibitem[Vilhena et~al.(2014)Vilhena, Foster, Rosvall, West, Evans, and
  Bergstrom]{vilhena2014finding}
Daril~A Vilhena, Jacob~G Foster, Martin Rosvall, Jevin~D West, James Evans, and
  Carl~T Bergstrom.
\newblock Finding cultural holes: How structure and culture diverge in networks
  of scholarly communication.
\newblock \emph{Sociological Science}, 1:\penalty0 221, 2014.

\bibitem[Wang et~al.(2021)Wang, Li, Naumann, Xiong, Cheng, Tinn, Wong, Usuyama,
  Rogahn, Shen, et~al.]{wang2021domain}
Yu~Wang, Jinchao Li, Tristan Naumann, Chenyan Xiong, Hao Cheng, Robert Tinn,
  Cliff Wong, Naoto Usuyama, Richard Rogahn, Zhihong Shen, et~al.
\newblock Domain-specific pretraining for vertical search: Case study on
  biomedical literature.
\newblock In \emph{Proceedings of the 27th ACM SIGKDD Conference on Knowledge
  Discovery \& Data Mining}, pages 3717--3725, 2021.

\bibitem[Weld and Bansal(2019)]{weld2019challenge}
Daniel~S Weld and Gagan Bansal.
\newblock The challenge of crafting intelligible intelligence.
\newblock \emph{Communications of the ACM}, 62\penalty0 (6):\penalty0 70--79,
  2019.

\end{thebibliography}

\end{document}